\RequirePackage{etoolbox}
\csdef{input@path}{{style/}{graphics/}}
\documentclass[aoas,MSNbibl,nameyear,seceqn,dvips]{arximspdf}
\makeatletter
   \@ifpackageloaded{graphicx}{}{\usepackage{graphicx}}
\makeatother
\usepackage{multirow}
\usepackage{graphicx}
\usepackage{url,breakurl}

\doi{10.1214/14-AOAS778} 
\volume{9}
\issue{1}
\pubyear{2015}
\firstpage{69}
\lastpage{93}
\docsubty{FLA}


\begin{document}
\begin{frontmatter}

\title{Ensembling classification models based on phalanxes of variables with applications in drug~discovery\thanksref{T1}}
\runtitle{Ensembling based on phalanxes of variables}

\begin{aug}
\author[A]{\fnms{Jabed H.}~\snm{Tomal}\thanksref{m1,m2}\ead[label=e1]{jtomal@utstat.toronto.edu}\ead[label=u1,url]{www.utsc.utoronto.ca/cms}},
\author[B]{\fnms{William J.}~\snm{Welch}\corref{}\thanksref{m3}\ead[label=e2]{will@stat.ubc.ca}\ead[label=u2,url]{www.stat.ubc.ca}} \and
\author[B]{\fnms{Ruben H.}~\snm{Zamar}\thanksref{m3}\ead[label=e3]{ruben@stat.ubc.ca}}
\runauthor{J. H. Tomal,  W. J. Welch and R. H. Zamar}
\affiliation{University of Dhaka\thanksmark{m1}, University of Toronto\thanksmark{m2}
and University of British Columbia\thanksmark{m3}}
\address[A]{J. H. Tomal \\
Department of Computer \\
\quad and Mathematical Sciences\\
University of Toronto Scarborough\\
Toronto, Ontario M1C 1A4 \\
Canada\\
\printead{e1}\\
\printead{u1}}
\address[B]{W. J. Welch\\
R. H. Zamar\\
Department of Statistics\\
University of British Columbia\\
Vancouver, British Columbia V6T 1Z4\\
Canada\\
\printead{e2}\\
\phantom{E-mail:\ }\printead*{e3}\\
\printead{u2}}
\end{aug}
\thankstext{T1}{Research funded by NSERC of Canada.}

\received{\smonth{5} \syear{2014}}

\begin{abstract}
Statistical detection of a rare class of objects in a two-class classification problem
can pose several challenges.
Because the class of interest is rare in the training data,
there is relatively little information in the known class response labels for model building.
At the same time
the available explanatory variables are often moderately high dimensional.
In the four assays of our drug-discovery application,
compounds are active or not against a specific biological target,
such as lung cancer tumor cells,
and active compounds are rare.
Several sets of chemical descriptor variables from computational chemistry
are available to classify the active versus inactive class;
each can have up to thousands of variables characterizing molecular structure
of the compounds.
The statistical challenge is to make use of the richness of the explanatory variables
in the presence of scant response information.
Our algorithm divides the explanatory variables into subsets adaptively
and passes each subset to a base classifier.
The various base classifiers are then ensembled to produce one model
to rank new objects by their estimated probabilities of belonging to
the rare class of interest.
The essence of the algorithm is to choose the subsets
such that variables in the same group work well together;
we call such groups \emph{phalanxes}.
\end{abstract}

\begin{keyword}
\kwd{Clustering}
\kwd{model selection}
\kwd{quantitative structure activity relationship}
\kwd{random forest}
\kwd{ranking}
\kwd{rare class}
\end{keyword}
\end{frontmatter}

\section{Introduction}

Our goal is detection of rare chemical compounds that are active against a given biological target,
such as lung cancer cells or the HIV virus.
Statistical detection of rare events in a highly unbalanced two-class situation
occurs in a variety of other applications.
Detection of credit card fraud [\citet{BolHan2002}],
spam email  [\citet{HasTibFri2009}],
terrorism threats
and finding relevant documents in a Google search
are all examples of this problem.

In drug discovery, rare active compounds are sought in huge chemical libraries.
Our goal is to develop a quantitative structure activity relationship (QSAR)
model relating the probability of activity
to variables characterizing chemical structure
for use in ranking a large number of candidate compounds
and produce a shortlist rich in active compounds.

For improved analysis of four such drug-discovery problems relating to four assays,
we propose a new classification methodology.
The response variable in each study is the $0/1$ compound activity status
against a specific biological target.
For each problem five descriptor sets of explanatory variables
are available to build a classifier of activity.
The variables in the descriptor sets characterize the chemical/molecular structures
of the compounds in different ways.
Some have thousands of variables.
In contrast, the assay data are relatively uninformative.
While the training data have thousands of compounds,
very few are active: the fraction of actives varies from about 1\% to 8\%
for the four assays.
Thus, this paper aims to exploit the riches of up to thousands of explanatory variables
in the descriptor sets
in the presence of limited response information caused by imbalance.

Recursive partitioning [\citet{HawKas1982}] and classification trees [\citet{BreFriOls1984}]
have been successful for modeling drug-discovery data.
\citet{RusFarLam1999} were able to apply recursive partitioning to large structure-activity data sets
with thousands to millions of molecular descriptors
by making recursive partitioning scale well computationally.

Ensemble methods that combine several classifiers to produce one model
are widely viewed as even more competitive for drug-discovery data.
The method of random forests [RF, \citet{Bre2001a}],
an ensemble of classification trees,
has attracted particular attention.
\citet{SveLiaTon2003,CheLiaBre2004} and \citet{PolMurArt2009}, for example,
all showed RF is a relatively accurate method for classifying chemical compounds
in QSAR studies.
\citet{BruMelPic2007} compared several machine learning tools for mining drug discovery data,
including support vector machines and ensembles based on classification trees:
bagging [\citet{Bre1996}], boosting [\citet{FreSch1996}] and RF.
The authors demonstrated that ensembles provide better predictive performances
than nonensemble methods.
\citet{HugBroWel2012} carried out a comprehensive comparison of 12 classifiers, including RF,
to rank the compounds in several highly unbalanced two-class assay data sets in QSAR studies.
Repeatedly, RF emerged as one of the best ranking procedures.
As we shall see,
even RF, which is one of the most competitive existing methods,
may only find a minority of the active compounds.
There is much room for improvement.

Ensemble methods such as bagging and RF create a number of models
to average by repeatedly perturbing the data.
RF also randomly selects the set of explanatory variables
considered at each iteration as a constituent tree is built.
In principle, however, like bagging, it has all variables available for each tree in the ensemble.
In contrast, the method we introduce ensembles classifiers built
with distinct subsets of variables.
The algorithm identifies a number of such subsets,
where the variables in a subset work well together in the same model.
We call such subsets \emph{phalanxes}.

This notion of phalanxes exploits the richness of the dimensionality of the
explanatory variables in the following way.
Each phalanx is a relatively low-dimensional subset of variables,
so each variable has an opportunity to play a role in its model fit.
In this way, variables in different models can contribute to the overall
classification model,
without competing with each other in the sense that one variable
deselects another.
Our phalanx-forming algorithm can also be thought of as a special type of clustering of variables, where ``similarity'' between a pair of variables or a pair of groups
is working well together in the same model,
and ``dissimilarity'' means working well
when separated in different models, which are ultimately ensembled.

Natural subsets of variables are sometimes suggested by subject matter knowledge.
For example, in a single nucleotide polymorphism (SNP) genotyping application,
 \citet{Po06} used pairs of variables suggested a priori
by the different chemical procedures employed in the genotyping platform.
In the group LASSO and its variants [\citet{YuaLin2006,MeiVanBuh2008}],
given groups are evaluated by their ability to work together with other groups,
but in a \emph{single} model.
Most related to the proposed method is ensembles over classifiers based on single variables,
pairs of variables, etc., as used in the thesis of \citet{Wan2005},
but again groups were not formed in a data-adaptive way.

In principle, any given base classification method can be used to model the class response variable as a function of the explanatory variables in a phalanx
and to guide the data-adaptive grouping into phalanxes.
We use RF here because of its documented competitive performance
for drug-discovery data.
Thus, our ultimate classifier is an ensemble across random forests,
itself an ensemble method, where each random forest only uses the variables in one
phalanx.

The remainder of the article is organized as follows.
Section~\ref{sect:data} describes the four
assay data sets and the five descriptor sets,
and Section~\ref{sect:performance} defines the assessment metrics to assess
classification performance in the context of ranking
for this application.
Section~\ref{sect:algorithm} describes the algorithm for phalanx formation,
leading to the final ensemble, which we call an ensemble of phalanxes classifier.
Section~\ref{sect:results} presents performance results and comparisons.
Comparisons are made with RF and regularized random forests [RRF, \citet{DenRun2013}],
and with methods specifically designed for imbalanced drug-discovery data.
For more than a few tens of explanatory variables,
the proposed method needs to form initial groups,
and results are also provided for an application-specific approach to grouping
versus more general, data-adaptive ways.
Section~\ref{sect:results} also explores the statistical diversity
of ensembles of phalanxes and the implications for finding chemically diverse
active compounds in the application.
Finally, Section~\ref{sect:conclusions} draws some conclusions.

\section{Data sets and variables}
\label{sect:data}

We analyze 20 data sets from four different assays in the Molecular Libraries Screening Center Network.
The response data can be downloaded from
\url{http://pubchem.ncbi.nlm.nih.gov/}.

For each assay the response variable is $y$,
where $y = 0,1$ denotes inactivity and activity, respectively,
of a compound against a specific biological target.
Table~\ref{tab:assays} summarizes the four assays.
\begin{table}
\tablewidth=\textwidth
\caption{Four assays from the Molecular Libraries Screening Center Network}
  \label{tab:assays}
\begin{tabular*}{\tablewidth}{@{\extracolsep{\fill}}llcc@{}}
\hline
               &                            & \textbf{Compounds} & \textbf{Proportion}\\
\textbf{Assay} & \textbf{Biological target} & \textbf{(active)} & \textbf{active}\\
\hline
AID 348 & Gaucher's disease & 4946~(48)\phantom{0} & 0.0097\\
AID 362 & Tissue-damaging leukocytes & 4279~(60)\phantom{0} & 0.0140 \\
AID 364 & Cytotoxicity      & 3311~(50)\phantom{0}  & 0.0151 \\
AID 371 & Lung tumor cells  & 3312~(278) & 0.0839 \\
\hline
\end{tabular*}
\end{table}
Further information about AID 348 may be obtained from
\url{https://pubchem.ncbi.nlm.nih.gov/assay/assay.cgi?aid=348},
and similarly for the other three assays.

These four assays were investigated by \citet{HugBroWel2012},
and all are imbalanced with a sparse proportion of active compounds.
For three of the assays the proportion of actives is around $0.01$;
these three have only 48--60 active compounds each,
posing difficulties for any statistical modeling method.
For example, a classification tree would soon run short of active-class objects
and tend to build a shallow tree, only using a few important variables.
If there are many important variables, some must be omitted
in the path to any one terminal node.

The assays cover a range of drug-discovery applications.
AID 348 screens for inhibitors of mutant forms of beta-glucocerebrosidase,
implicated in Gaucher's disease.
AID 362 is a whole-cell assay for another
inhibitor of peptide binding,
associated with  tissue-damaging chronic inflammation.
Multiple mechanisms of activity [\citet{YouHaw1998}] are possible
even with the specific biological targets of AID 348 and 362
and even more likely for the other two assays, AID 364 and AID 371.
They are whole-cell live/dead assays.
Multiple mechanisms of activity, from multiple chemical structures,
call for correspondingly broad statistical modeling strategies
such as the ensemble method proposed herein.
In our approach,
constituent statistical models use distinct sets of explanatory
variables.

The principle underlying QSAR modeling in drug discovery is that
activity (toxicity/drug potency) of a chemical compound is related to its molecular structure,
which can be characterized by chemical descriptors.
These explanatory variables or covariates are
numeric variables that describe the structure or shape of molecules.

We consider five sets of descriptors for each of the four assays,
to give a total of $4 \times 5 = 20$ data sets.
The descriptor sets are the following: atom pairs (AP);
Burden numbers (BN) [\citet{Bur1989,PeaSmi1999}];
Carhart atom pairs (CAP) [\citet{CarSmiVen1985}]; fragment pairs (FP);
and pharmacophores fingerprints (PH).
The Burden numbers are continuous descriptors,
and the other four are bit strings where each bit is set to
``1'' when a certain feature is present and ``0'' when it is not.
See \citet{LiuFenYou2005} and \citet{HugBroWel2012} for further explanation
of the molecular properties captured by the descriptor sets.

Table~\ref{tab:descriptors} summarizes the five descriptor sets,
as generated by PowerMV [\citet{LiuFenYou2005}].
\begin{table}
\tablewidth=\textwidth
\caption{Five descriptor sets generated by PowerMV
and the number of nonconstant variables for each of the four assays}
\label{tab:descriptors}
\begin{tabular*}{\tablewidth}{@{\extracolsep{\fill}}lccccc@{}}
\hline
& & \multicolumn{4}{c@{}}{\textbf{Variables for assay}} \\[-4pt]
& & \multicolumn{4}{l@{}}{\hrulefill}\\
\textbf{Descriptor set} &  \multicolumn{1}{c}{\multirow{2}{36pt}[11pt]{\textbf{Potential} \textbf{variables}}} & \textbf{AID 348} & \textbf{AID 362}
& \textbf{AID 364} & \textbf{AID 371} \\
\hline
Atom pairs (AP)           &  \phantom{0}546 & \phantom{0}367 & \phantom{0}360 & \phantom{0}380 & \phantom{0}382\\
Burden numbers (BN)       &   \phantom{00}24 & \phantom{00}24 & \phantom{00}24 & \phantom{00}24 & \phantom{00}24\\
Carhart atom pairs (CAP)  & 4662 & 1795 & 1319 & 1585 & 1498\\
Fragment pairs (FP)       &  \phantom{0}735 & \phantom{0}570 & \phantom{0}563 & \phantom{0}580 & \phantom{0}580\\
Pharmacophores (PH)       &  \phantom{0}147 & \phantom{0}122 & \phantom{0}112 & \phantom{0}120 & \phantom{0}119\\
\hline
\end{tabular*}\vspace*{-6pt}
\end{table}
PowerMV computes a total of 546, 24, 4662, 735 and 147 descriptor variables
for AP, BN, CAP, FP and PH, respectively.
For the molecules in any given assay (see Table~\ref{tab:assays}),
some of the descriptors may be constant (e.g., a chemical feature is always absent).
Such constant variables are removed,
giving the numbers of nonconstant variables in Table~\ref{tab:descriptors}.

The bit-string descriptors have hundreds to thousands of variables,
with CAP having the most (1319--1795).
The continuous BN descriptors have the lowest dimensionality, with 24 variables for all assays.
They are also rich, however, in the sense that continuous variables possess good resolution
compared to binary variables.

\section{Performance measures}
\label{sect:performance}

We describe assessment metrics specific to evaluating ranking procedures
when the goal is to detect the few instances of the rare class
hidden in a large set of objects,
as in finding active compounds in a large chemical library.
These metrics are used instead of
misclassification error, a standard criterion
for classification performance in general,
but one that is inappropriate for highly unbalanced classes [\citet{ZhuSuChi2006}].

For a given classifier,
ranking of the compounds in a test set is based on
their estimated probabilities of activity, $\hat{\pi}$.
The compound with the largest $\hat{\pi}$ is ranked first, etc.
The goal is to rank the actives in the test set at the top of the list.
The performance measures relate to where the actives are in the ranked list.

Let $N$  be the total number of compounds in a test set,
and let $M \le N$ be the number of actives among them.
Suppose the ranked list is cut off at $n$ compounds;
for example, resources only allow follow-up of $n$ leads from the list.
Let $0\leq H(n) \leq M$ be the number of actives or ``hits'' in the shortlist of size $n$.
Performance is measured by graphing $H(n)$
or by computing one or more numerical functionals of it.

\subsection{Hit curve}
The hit curve is a plot of $H(n)$ versus $n$
or, equivalently, a~plot of $H(n)/M$ (proportion of actives found)
versus $p(n)=n/N$ (proportion of test compounds considered).
The hit curve shows the ranking performance at all possible shortlist cutoff-points, $n$.
Classifier $1$ with hit curve $H_1(n)$ is uniformly superior to
classifier $2$ with hit curve $H_2(n)$ if $H_1(n) \geq H_2(n)$ for $1 \le n \le N$,
with strict inequality for at least one value of $n$.

\begin{figure}

\includegraphics{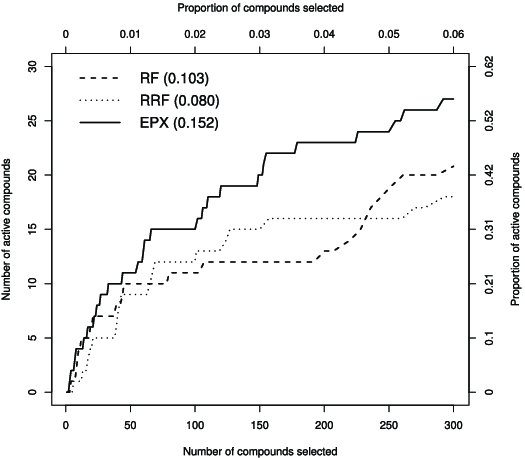}

\caption{Hit curves from three classifiers, RF, RRF, and EPX,
for the AID 348 assay and BN descriptors.
The numbers in the legend are values of AveP, defined below in Section~\protect\ref{sect:AveP}.}
\label{fig:hitcurves:aid348:BN}
\end{figure}

For example,
Figure~\ref{fig:hitcurves:aid348:BN} shows hit curves for three ensemble classifiers:
RF applied to all the available variables,
RRF,
and ensemble of phalanxes (EPX, described in Section~\ref{sect:algorithm}).
The three classifiers are applied here to the AID 348 assay and the BN descriptors.
The plots show $n$ up to $300$,
because actives are sought early in a ranked list.
Note that EPX dominates the other two ensembles:
its hit curve is uniformly above the other two.
There is no clear winner between RRF and RF,
because their hit curves cross.

In the results of Section~\ref{sect:results} we compare numerous hit curves,
and it is convenient to have a numerical criterion to summarize a hit curve.
Two such criteria are outlined next.

\subsection{Average precision}  \label{sect:AveP}
The average precision (AveP) gives a single number summary for a hit curve.
Suppose we shortlist the top $n \leq N$ compounds and $H(n)$ of them are active. Then
\[
h(n) =\frac{H(n)}{n} \in [0,1]
\]
is the hit rate or precision for the top $n$ ranked compounds.
Naturally, we want $h(n)$ to be as large as possible at every $n$.
Let $1\leq t_1 < t_2 < \cdots < t_M \leq N$ be the positions
of the $M$ active compounds in the ranked list.
AveP is defined as the average of the hit rates at the points on the hit curve where actives are found:
\[
\mbox{AveP} = \frac{1}{M} \bigl[ h(t_1) + h(t_2) +
\cdots + h(t_M)\bigr].
\]
AveP reaches the maximum value $1$ when all of the actives are ranked before all the inactives.
When there are tied $\hat{\pi}$ values, and hence ties in the ranked list of compounds,
the expected value of AveP is computed under random ordering of the compounds within each
group of ties [\citet{Wan2005}, Chapter~3].

The AveP values for RF, RRF and EPX in Figure~\ref{fig:hitcurves:aid348:BN} are $0.103$, $0.080$ and $0.152$, respectively.
EPX is a clear winner by the AveP measure,
which makes sense as its hit curve dominates the other two.
By the AveP criterion, RF is preferable to RRF.

We use AveP not only to evaluate classifiers
but also to choose the phalanxes in our algorithm (Section~\ref{sect:algorithm}).

\subsection{Initial enhancement}
Initial enhancement (IE), defined by \citet{KeaSalFlu1996},
is the precision at one specific shortlist length, $n$,
normalized by the proportion of actives in the entire collection of compounds:
\[
\mbox{IE} = \frac{h(n)}{M/N}.
\]
Because IE is just a rescaling of the precision at $n$,
both measures would lead to the same conclusions,
but IE is often given in QSAR studies
to measure the improvement relative to the expectation under random ranking.
Naturally, IE values (much) larger than 1 are desired.
A drawback of IE is that it depends on the particular shortlist size, $n$.
Moreover, IE does not distinguish whether the actives are ranked
at the beginning or end of the shortlist.
Therefore, while we report IE performance results,
the AveP criterion is used to choose phalanxes in Section~\ref{sect:algorithm}.

Following \citet{HugBroWel2012}, we use $n=300$ throughout to calculate IE.
The IE values for RF, RRF and EPX in Figure~\ref{fig:hitcurves:aid348:BN}
are $7.15$, $6.18$ and $9.27$, respectively.
Again EPX is the winner, and RF is the runner-up.

\subsection{Balanced 10-fold cross-validation (CV)}

The assay data summarized in Table~\ref{tab:assays} are used
for training classifiers and testing them.
Because actives are sparse,
throughout we use balanced 10-fold cross-validation to assess performance.
Thus, we randomly divide the data into $10$ approximately equal sized groups,
each containing approximately $1/10$ of the actives.
When a group serves as a test set,
and the remaining nine groups are the training data for a classifier,
a~$\hat{\pi}$ value is obtained for every compound in the test set.
After all 10 groups have served as a test set,
$\hat{\pi}$ values are available for all compounds and  can be ranked
to give a hit curve, as in Figure~\ref{fig:hitcurves:aid348:BN},\vadjust{\goodbreak} or to compute AveP or IE.

\section{Phalanx-formation algorithm}
\label{sect:algorithm}

\subsection{Phalanxes of variables}

We borrow the term \emph{phalanx} from the military formation
used by Alexander The Great and his father Philip II of Macedon
to deploy infantry soldiers in the battlefield.
For psychological motivation,
phalanxes were organized as groups of friends and family members.
As a result,
the strength of a phalanx would depend upon the individual strengths of its soldiers
and the emotional bonds between them.
A phalanx was an autonomous fighting unit but could be ensembled with other phalanxes
to form a formidable military machine.

The analogy with classification is that the proposed algorithm
selects a group of variables in a statistical phalanx
such that they form a strong classifier when put together in a single model.
In other words, the variables in a phalanx work better together in a model
than when separated in different models.
At the same time, the algorithm pays attention to the performance
of the overall strength of the final ensemble of models.

\subsection{Phalanx formation}
\label{sect:phalanx:formation}

Even if the optimal number of phalanxes is known,
dividing the variables into phalanxes is a combinatorial problem.
With the higher-dimensional descriptor variables in Table~\ref{tab:descriptors},
exhaustive search is infeasible,
and the algorithm performs a greedy (look one iteration ahead) optimization instead.
The amalgamation of variables into phalanxes resembles hierarchical clustering,
but variables are clustered, not observations.

\begin{figure}

\includegraphics{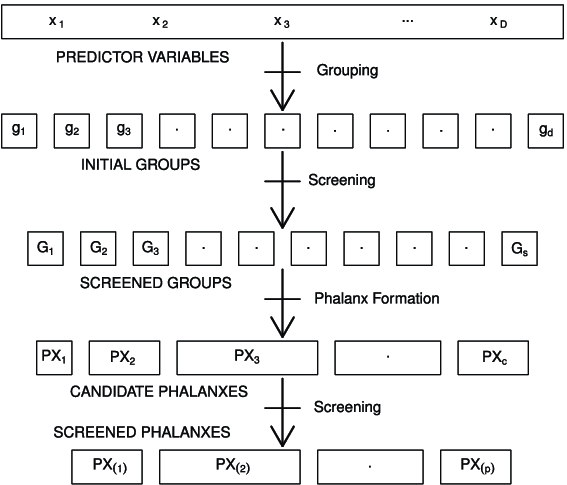}

\caption{Algorithm for phalanx formation.
$D$ variables are partitioned into $d$ initial groups,
screened down to $s$ groups,
combined into $c$ candidate phalanxes,
and then screened down to $p$ phalanxes in the final ensemble
$(D \geq d \geq s \geq c \geq p)$.}
\label{phalanxformation}
\end{figure}

As shown in Figure~\ref{phalanxformation},
there are four main steps in the algorithm to group
the original $D$ variables into $p$ final phalanxes:
\begin{longlist}[4.]
\item[1.] Initial grouping.
The original $D$ variables are partitioned into $d \le D$ initial groups.
\item[2.]
Screening.
The $d$ initial groups are screened down to $s \le d$ groups.
\item[3.]
Hierarchical merging into phalanxes.
The $s$ screened groups are amalgamated hierarchically into $c \le s$ candidate phalanxes.
\item[4.]
Screening.
The $c$ candidate phalanxes are screened down to $p \le c$ final phalanxes.
\end{longlist}

At termination, $p$ base classifiers are trained, one for each phalanx of variables.
They form an ensemble of models:
the $\hat{\pi}$ values from the $p$ classifiers
are averaged to give one value of $\hat{\pi}$ for each compound.

We use RF as the base classifier for the ensemble of phalanxes throughout,
including phalanx formation.
There are two reasons.
As already mentioned, RF is known to be a competitive method for drug-discovery data,
so it will provide strong base classifiers for our ensemble.
Second, during phalanx formation its out-of-bag (OOB) estimated class probabilities
provide as good an assessment of performance as cross-validation
[\citeauthor{Bre1996b} (\citeyear{Bre1996b,Bre2001a}),  \citet{Tib1996,WolMac1999}].
Thus, the computational expense of multiple fits in cross-validation is avoided.
To further reduce computation, for phalanx formation the number of trees grown
is reduced from the default of 500 to 150.

All steps in phalanx formation are guided by a chosen assessment measure, $a$.
We use AveP for $a$,
but the final ensembles are also evaluated using IE,
and the algorithm trivially generalizes to other criteria.
(The algorithm assumes $a$ is to be maximized,
but the changes for a smaller-the-better criterion are straightforward.)

The four steps of the algorithm are now described in greater detail;
\begin{longlist}[4.]
\item[1.] \textit{Initial grouping.}
This optional step has two motivations.
First, the dimensionality of the AP, CAP, FP and PH descriptors
in Table~\ref{tab:descriptors}
makes the later hierarchical-merging step too computationally expensive,
even with the greedy implementation.
At each iteration the algorithm considers amalgamating all pairs of groups,
and the computational complexity (see Section~\ref{sect:comp:complex})
is still quadratic in the number of groups,
too demanding for all but the BN descriptors.
Second, the four higher-dimensional sets have binary variables,
and a single binary variable can only give two possible $\hat{\pi}$ values.
Consequently, the initial classifiers are all extremely weak.
In contrast, an initial group of $k > 1$ binary variables can generate up to $2^k$
possible ranks.

Guidance is available to group the AP, CAP, FP or PH descriptors
from the variable names.
For FP, for example, there are seven variables
relating to the presence of two aromatic rings:
$\mbox{AR}\_01\_\mbox{AR}, \mbox{AR}\_02\_\mbox{AR},
\ldots, \mbox{AR}\_07\_\mbox{AR}$.
Here, $\mbox{AR}\_01\_\mbox{AR}$ represents two phenyl (aromatic) rings separated by one bond, etc.
These seven variables form one of the initial groups,
similarly the other groups.
We have empirically verified that grouping by the variable names
provides better final classification accuracy than forming initial groups at random.
A data-adaptive alternative for grouping is presented in Section~\ref{sect:igroups}.

In this way,
the $D$ original variables are grouped into $d\leq D$ initial groups,
denoted by $g_1,g_2,\ldots,g_d$.
For the BN descriptors, this step is omitted:
the initial groups are the individual variables,
that is, $d = D = 24$.

\item[2.] \textit{Screening of initial groups}.
We screen out weak initial groups to reduce computational burden and noise.

To survive this step, a group must be strong in the sense of
a comparison with the distribution of the assessment criterion, $a$,
under random ranking.
To compute this reference distribution,
we randomly permute the $0/1$ values of the response variable $y$ relative to
the descriptor values
and obtain the corresponding value of~$a$.
Repeating for many random permutations results in an empirical distribution of $a$
(we use 1000 repeats throughout),
from which we take the $\alpha$ quantile,
denoted by~$a_\alpha$ (we use $\alpha = 0.95$ for all reported results).
The algorithm also makes use of $a_{0.5}$,
the median of the empirical distribution.

An initial group is deemed to be strong if the base classifier using its variables
is competitive with $a_\alpha$.
There are actually three tests, and
a group $g_i$ survives the initial screening if it passes at least one of them.
The three tests consider the performance of $g_i$
by itself, or when its variables are combined with those in any other group $g_j$,
or when it forms an ensemble with any other group.

Thus, we need to define the following performance measures.
Denote by $\hat{\pi}(g_i)$ the estimated probabilities of activity from
the base classifier using only the variables in $g_i$,
and let $a_i=a (\hat{\pi}(g_i) )$ be
the assessment measure.
Similarly,
denote by $\hat{\pi}(g_i \cup g_j)$ the estimated probabilities of activity when
the variables in $g_i$ {and} $g_j$ ($i \ne j$)
are all available to the base classifier to fit a single model,
and let
%
\begin{equation}\label{eqn:aij}
a_{ij} = a \bigl(\hat{\pi}(g_i \cup g_j)
\bigr)
\end{equation}
be the resulting performance measure.
Finally,
consider the performance of an ensemble of two models based on $g_i$ and $g_j$,
respectively.
Probability averaging of their two sets of estimated probabilities gives
$ (\hat{\pi}(g_i) + \hat{\pi}(g_j) )/2$ for ranking.
The resulting assessment measure is
%
\begin{equation}\label{eqn:aij:bar}
a_{\overline{ij}} =a \bigl( \bigl(\hat{\pi}(g_i)+\hat{
\pi}(g_j) \bigr)/2 \bigr).
\end{equation}

Based on these various uses of $g_i$ and the corresponding assessment measures,
$g_i$ is deemed to be strong and survives the initial screening if
it passes at least one of the following tests:
\begin{itemize}
\item
$g_i$ is strong alone:
%
\begin{equation}\label{eqn:strong1}
a_i \geq a_\alpha.
\end{equation}
\item $g_i$ improves the strength of another group $g_j$
when $g_i$ and $g_j$ are used together in a single model:
%
\begin{equation}\label{eqn:strong2}
a_{0.5} + a_{ij} - a_{j} \geq a_\alpha\qquad \mbox{for at least one $j \ne i$}.
\end{equation}
The rationale is that $a_{ij} - a_{j}$ is the improvement from
adding the variables in $g_i$ to those in $g_j$ in a single model,
an improvement that has to be competitive with $a_\alpha - a_{0.5}$.
\item
$g_i$ improves the strength of another group $g_j$
when $g_i$ and $g_j$ are in an ensemble of two models:
%
\begin{equation}\label{eqn:strong3}
a_{0.5} + a_{\overline{ij}} - a_{j} \geq a_\alpha
\qquad\mbox{for at least one $j \ne i$}.
\end{equation}
\end{itemize}

After removing weak initial groups,
the list of surviving groups is relabeled as
$\{G_1,G_2,\ldots,G_s\}$ for the next step.

\item[3.] \textit{Hierarchical merging into phalanxes.}
This step to merge $G_1, G_2,\ldots,G_s$ into phalanxes of variables
is the heart of the algorithm.
It resembles hierarchical clustering, but merges groups of variables,
not groups of observations.

Each iteration merges the pair of groups $G_i$ and $G_j$ that minimizes
\[
m_{ij} = a_{\overline{ij}}/a_{ij},
\]
where $a_{\overline{ij}}$ and $a_{ij}$ are defined in~(\ref{eqn:aij:bar}) and~(\ref{eqn:aij}).
Values of the ratio less than 1~indicate that $G_i$ and $G_j$ perform better in a
single model than when ensembled in separate models.
After each merge, the number of groups, $s$, is reduced by 1,
and one of the new groups is the union of two of the old groups.
The algorithm continues until $m_{ij} \geq 1$ for all $i,j$,
suggesting that merging reduces performance and the groups
should be ensembled.

The following example illustrates.
For simplicity, consider only $s=3$
initial groups (actually individual variables)
from the BN descriptors and\break assay \mbox{AID 348}.
The three groups are $G_1 = \mbox{WBN}\_\mbox{GC}\_\mbox{L}\_1.00$,
$G_2 = \break \mbox{WBN}\_\mbox{EN}\_\mbox{H}\_0.50$ and
$G_3 = \mbox{WBN}\_\mbox{LP}\_\mbox{H}\_1.00$.
The AveP values when pairs of groups are used together in a single model are
$a_{12}=0.052$, $a_{13}=0.037$ and $a_{23}=0.054$.
When pairs of groups are ensembled,
the AveP values are $a_{\overline{12}}=0.069$, $a_{\overline{13}}=0.050$ and $a_{\overline{23}}=0.031$.
Thus, the corresponding $m_{ij}$ ratios are $1.31$, $1.36$ and $0.57$.
As the variables ${G}_2$ and ${G}_3$ give the smallest $m_{ij}$
and it is less than $1$,
we merge ${G}_2$ and ${G}_3$ into a new group and there are now $s=2$ groups.
At the next step it turns out that $m_{12} = 1.18$ and the two new groups
should not be merged.
Thus, the algorithm terminates with two candidate groups or phalanxes,
one of which contains two of the original variables.

In general, the $c$ final groups are candidate phalanxes,
$\mbox{PX}_1, \mbox{PX}_2, \ldots, \allowbreak \mbox{PX}_c$.

\item[4.] \textit{Screening out weak phalanxes.}
A candidate phalanx is kept in the ensemble if it is
individually strong as defined in~(\ref{eqn:strong1})
or it is strong in an ensemble with another phalanx as defined in~(\ref{eqn:strong3}).
There is no need to check condition~(\ref{eqn:strong2}),
as there was an exhaustive search for merging groups in the previous step.

The $p$ surviving phalanxes from this second stage of screening
form the army or ensemble, $\mbox{PX}_{(1)},\ldots,\mbox{PX}_{(p)}$,
for ranking.
\end{longlist}

\subsection{Ensemble of phalanxes}
\label{sect:epx}
We fit $p$ RF classifiers,
one for each of the $p$ phalanxes of variables,
and obtain probabilities of activity from them.
Here, 500 trees are grown for each random forest, the default.
For any test point, the $p$ probabilities of activity from the ensemble of phalanxes (EPX)
are averaged to give the final probabilities for ranking.

\subsection{Computational complexity}
\label{sect:comp:complex}

We now show that the computational complexity of phalanx formation
is $\mathcal{O} (d^2)$ fits of the underlying base classifier in the worst case.
Recall $d$ is the number of initial groups,
or the number of variables if there is no grouping.

The screening phase first involves $d$ fits, one for each group.
Then, models are fit for all the unions of all possible pairs of groups,
that is, $d(d - 1) /2$ fits.
Hence, screening involves a total of $d(d + 1) /2$ fits.
In the worst case, no groups are removed by screening.

For the first merger of the phalanx formation stage,
no new fits are required:
the performance measures for individual groups and pairs of groups
to evaluate~(\ref{eqn:aij})
are already available from the screening stage.
If the algorithm continues,
two groups are merged to create a new one.
At the second iteration,
one fit needs to be made for the new group just formed
as well as another $d - 2$ fits using the union of variables
from the new group and one of the other $d - 2$ groups,
that is, a total of $d - 1$ fits.
Note that we do not need to refit models for all possible pairs of groups.
In the next iteration, with one fewer group,
there are $d - 2$ fits, etc.
In the worst case, phalanx formation continues until there is
only one phalanx remaining,
for a total of $(d - 1) + (d - 2) + \cdots + 1 = d (d - 1) / 2$ fits.

Thus, between screening and phalanx formation,
there are at most $d^2$ fits.
The computational burden caused by the dimensionality
of four of the five descriptor sets is greatly reduced by forming initial groups
with $d \ll D$.
Moreover, the computations are embarrassingly parallel:
The many fits necessary at any iteration can be made independently of each other.
Thus, parallel computation is straightforward using the R packages
\texttt{foreach}, \texttt{iterators}, \texttt{doSNOW} and \texttt{doMPI}.

\section{Results}
\label{sect:results}

\subsection{Comparison with random forests}

We consider 20 data sets: there are four assays and each has five possible descriptor sets.
For each data set an EPX classifier is constructed,
as described in Section~\ref{sect:algorithm};
this is repeated three times using different random seeds.
The results from EPX are compared with RF and RRF,
constructed using the defaults in their respective R packages,
\texttt{randomForest} [\citet{LiaWie2002}] and \texttt{RRF} [\citet{DenRun2013}].
We give detailed results for AID 348.
Summary results will also be given
for the other three assays (AID 362, AID 364 and AID 371).

The last five columns of Table~\ref{tab:phalanxes:aid348}
summarize the steps in the EPX algorithm of Section~\ref{sect:algorithm} for AID 348.
For example, the descriptor set AP has a total of $367$ variables
arranged into $75$ initial groups, of which $22$ survive screening in the first EPX run.
The $22$ groups are ultimately merged into $4$ candidate phalanxes,
of which $2$ survive screening.
Thus, the final EPX classifier is an ensemble of 2 phalanxes.
Two further runs of EPX with different random seeds
result in armies of 5 and 4 phalanxes, respectively.
(The impact of the variation in the number of phalanxes is reported later in this section.)
\begin{table}
\tablewidth=\textwidth
\caption{Number of variables, initial groups, screened groups, candidate phalanxes
and screened phalanxes for the AID 348 assay and five descriptor sets.
There are three runs of the EPX algorithm}
\label{tab:phalanxes:aid348}
\begin{tabular*}{\tablewidth}{@{\extracolsep{\fill}}lcccccc@{}}
\hline
& & \multicolumn{5}{c@{}}{\textbf{Number} \textbf{of}}\\[-4pt]
&& \multicolumn{5}{l@{}}{\hrulefill}\\
 \multirow{2}{43pt}[-5pt]{\textbf{Descriptor} \textbf{set}} & & & \multicolumn{2}{c}{\textbf{Groups}} & \multicolumn{2}{c@{}}{\textbf{Phalanxes}}\\[-6pt]
&&& \multicolumn{2}{l}{\hrulefill} & \multicolumn{2}{l@{}}{\hrulefill}\\
 & \textbf{Run} & \textbf{Variables} & \textbf{Initial} & \textbf{Screened} & \textbf{Candidate} & \textbf{Screened} \\
\hline
AP  & $1$   & \phantom{0}367  & \phantom{0}75   & \phantom{0}$22$ & \phantom{0}$4$ & \phantom{0}$2$  \\
    & $2$   &
 &
 & \phantom{0}$19$  & \phantom{0}$8$  & \phantom{0}$5$        \\
    & $3$   &
  &
 & \phantom{0}$22$  & \phantom{0}$8$  & \phantom{0}$4$      \\[3pt]
BN  & $1$   & \phantom{00}24   & \phantom{0}24   & \phantom{0}$24$ & \phantom{0}$8$ & \phantom{0}$8$  \\
    & $2$   &
 &
  & \phantom{0}$24$  & \phantom{0}$9$  & \phantom{0}$9$  \\
    & $3$   &
  &
 & \phantom{0}$24$  & \phantom{0}$4$  & \phantom{0}$4$       \\[3pt]
CAP & $1$   & 1795 & 455  & 398  & 13  & $10$ \\
    & $2$   &
  &
  & $128$ & \phantom{0}$8$  & \phantom{0}$8$     \\
    & $3$   &
  &
  & $352$ & $17$ & $12$     \\[3pt]
FP  & $1$   & \phantom{0}570  & 101  & \phantom{0}24   & \phantom{0}6   & \phantom{0}4    \\
    & $2$   &
  &
  & \phantom{0}$22$  & \phantom{0}$5$  & \phantom{0}$4$      \\
    & $3$   &
  &
  & \phantom{0}$22$  & \phantom{0}$5$  & \phantom{0}$5$      \\[3pt]
PH  & $1$   & \phantom{0}120  & \phantom{0}$21$ & \phantom{00}$5$  & \phantom{0}$1$ & \phantom{0}$1$  \\
    & $2$   &
  &
  & \phantom{00}$5$   & \phantom{0}$3$  & \phantom{0}$2$     \\
    & $3$   &
  &
  & \phantom{00}$5$   & \phantom{0}$1$  & \phantom{0}$1$      \\
\hline
\end{tabular*}
\end{table}

Table~\ref{tab:phalanxes:aid348} shows that many of the initial groups
are screened out from the four descriptor sets based on binary variables: AP, CAP, FP, and PH.
For example, 71--75\% of AP's initial groups are dropped.
For the other binary descriptor sets, 13--72\%, 76--78\% and 76\% of the initial
groups are screened out.
In contrast, all of the continuous BN variables are always used.

The AveP performance measures are reported in Table~\ref{mean:AveP:aid348} for EPX, RF and RRF.
(The description of two further methods, SRF and WRF appearing in Table~\ref{mean:AveP:aid348},
is taken up in Section~\ref{sect:dataimbalance}.)
All results are based on balanced 10-fold cross-validation.
Because of randomness in cross-validation, especially with such small frequencies
of active-class compounds,
cross-validation is repeated $16$ times for different random, balanced data splits
(16 times because we initially had 16 processors conveniently available for parallel processing).
Table~\ref{mean:AveP:aid348} therefore reports mean AveP across the 16 cross-validations.
For the binary descriptor sets AP, CAP, FP and PH---where there are many descriptor variables
and screening is presumably important---RRF outperforms RF.
But for BN, RF outperforms RRF.
Mean AveP is always largest for EPX, however.
The advantage of EPX is greatest for CAP,
the set with the largest number of descriptors.

Columns 8 and 9 in Table~\ref{mean:AveP:aid348} show that
EPX consistently beats RF and RRF in all 16 repeats of cross-validation
across all descriptor sets.

\begin{table}
\tablewidth=\textwidth
\caption{AveP averaged over $16$ replications of balanced $10$-fold cross-validation
for EPX, RF, RRF, SRF and WRF applied to the AID 348 assay and five descriptor sets (DS).
The last four columns show the number of times EPX has larger AveP among the $16$ repeats
of cross-validation relative to RF, RRF, SRF and WRF}
\label{mean:AveP:aid348}
\begin{tabular*}{\tablewidth}{@{\extracolsep{\fill}}lcccccccccc@{}}
\hline
&  & \multicolumn{5}{c}{\textbf{Mean} \textbf{AveP}} & \multicolumn{4}{c@{}}{\textbf{EPX} \textbf{beats}} \\[-4pt]
&& \multicolumn{5}{l}{\hrulefill} & \multicolumn{4}{l@{}}{\hrulefill}\\
\textbf{DS} & \textbf{Run} & \textbf{EPX} & \textbf{RF} & \textbf{RRF} & \textbf{SRF} & \textbf{WRF} & \textbf{RF} & \textbf{RRF} & \textbf{SRF} & \textbf{WRF}\\
\hline
AP & 1 & $0.182$ & $0.063$  & $0.081$  & $0.052$  & $0.058$  & $16/16$ & $16/16$ & $16/16$ & $16/16$\\
& 2 & $0.194$ &
 & & & & $16/16$ & $16/16$ & $16/16$ & $16/16$ \\
& 3 & $0.146$ &
& &  & & $16/16$ & $16/16$ & $16/16$ & $16/16$ \\[3pt]
BN  & 1 & $0.143$ & $0.090$  & $0.078$  & $0.075$  & $0.075$  & $16/16$ & $16/16$ & $16/16$ & $16/16$  \\
& 2 & $0.153$ &
& &  & & $16/16$ & $16/16$ & $16/16$ & $16/16$ \\
& 3 & $0.132$ &
 & &  & & $16/16$ & $16/16$ & $16/16$ & $16/16$  \\[3pt]
CAP  & 1 &  $0.201$ & $0.068$  & $0.090$  & $0.095$ & $0.088$  & $16/16$ & $16/16$ & $16/16$ & $16/16$ \\
& 2 & $0.184$ &
& &  & & $16/16$ & $16/16$ & $16/16$ & $16/16$  \\
& 3 & $0.155$ &
 & &  & & $16/16$ & $16/16$ & $15/16$ & $16/16$ \\[3pt]
FP  & 1 & $0.157$ & $0.077$  & $0.098$  & $0.091$  & $0.099$  & $16/16$ & $16/16$ & $16/16$ & $16/16$  \\
& 2 & $0.130$ &
& &  & & $16/16$ & $16/16$ & $15/16$ & $16/16$  \\
& 3 & $0.157$ &
& &  & & $16/16$ & $16/16$ & $16/16$ & $16/16$  \\[3pt]
PH  & 1 & $0.108$ & $0.070$  & $0.080$  & $ 0.082$  & $0.060$  & $16/16$ & $16/16$ & $15/16$ & $16/16$  \\
& 2 & $0.108$ &
 & &  & & $16/16$ & $16/16$ & $15/16$ & $16/16$  \\
& 3 & $0.108$ &
& &  & & $16/16$ & $16/16$ & $15/16$ & $16/16$ \\
\hline
\end{tabular*}
\end{table}

\begin{figure}

\includegraphics{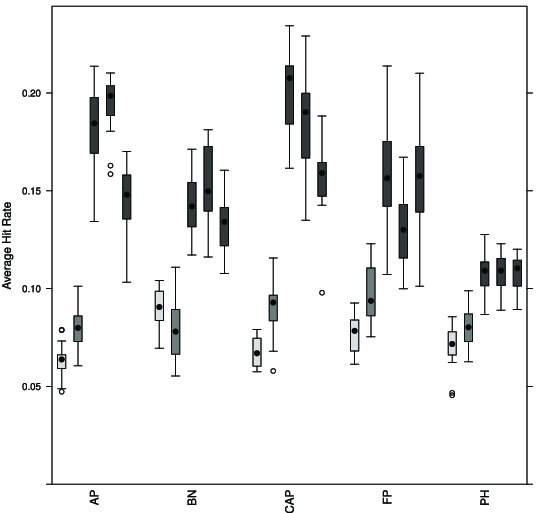}

\caption{Box-plots of AveP for the AID 348 assay and five descriptor sets
from $16$ replications of balanced cross-validation
for RF (light gray), RRF (gray)
and $3$ armies of phalanxes (dark gray).
}
\label{fig:AveP:aid348}
\end{figure}

Figure~\ref{fig:AveP:aid348} shows box-plots for the $16$ values of AveP over the 16 cross-validations.
The three box-plots for EPX correspond to the three runs---with different random seeds---in our cross-validation experiment.
Despite exhibiting some run-to-run variability,
EPX consistently outperforms RF and RRF.
We could stabilize the algorithm by using RF
with a larger number of trees at the phalanx-formation stage,
but this would increase the computational burden of the algorithm.

\begin{figure}
\begin{tabular}{@{}cc@{}}

\includegraphics{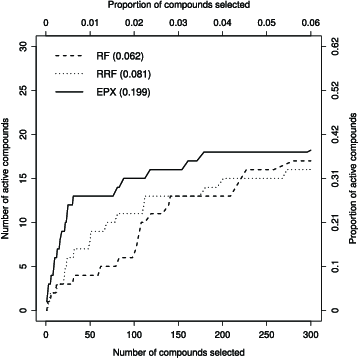}
 & \includegraphics{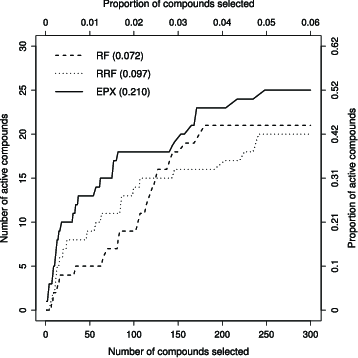}\\
\footnotesize{(a) Atom pairs (AP)} & \footnotesize{(b) Carhart atom pairs (CAP)}\\[6pt]

\includegraphics{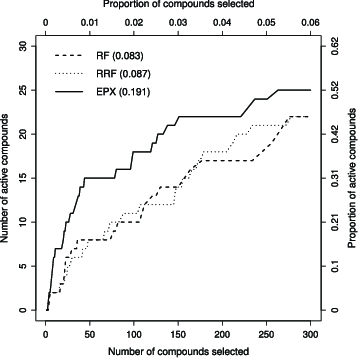}
 & \includegraphics{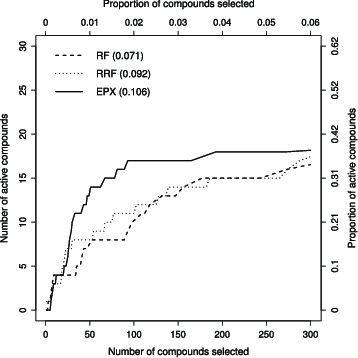}\\
\footnotesize{(c) Fragment pairs (FP)} & \footnotesize{(d) Pharmacophores (PH)}
\end{tabular}
 \caption{Hit curves from RF (dashed line),
   RRF (dotted line) and EPX (solid line) for the AID $348$ assay.
    Results are given for four descriptor sets: \textup{(a)} AP; \textup{(b)} CAP; \textup{(c)} FP; and \textup{(d)} PH.
     The numbers in parentheses in the legends are the values of AveP.}
        \label{fig:hitcurves:aid348}
\end{figure}

To visualize the performance gains of EPX,
the hit curves in Figure~\ref{fig:hitcurves:aid348:BN} (for descriptor set BN)
and Figure~\ref{fig:hitcurves:aid348} (for the other descriptor sets)
are from the first balanced 10-fold cross-validation.
In all cases the hit curve for EPX starts rising very quickly
and dominates the curves for the other two methods.
In other words, EPX is more successful here in detecting actives early in a list of ranked
compounds.

Although we have formed armies of phalanxes by optimizing AveP,
Table~\ref{tab:IE:aid348} shows they also have good performance in terms of the metric IE.
Mean IE averaged over the 16 replications of balanced cross-validation is
consistently larger for EPX.

EPX also exhibits strong performance when similar studies
are performed for the other three assays, AID 362, AID 364 and AID 371.
Summary performance measures are given in
Tables~\ref{tab:mean:AveP:aid362}, \ref{tab:mean:AveP:aid364} and \ref{tab:mean:AveP:aid371},
respectively.
For these assays, the performances of EPX and RF are comparable for the smallest
set of binary descriptors, PH, but EPX dominates for CAP,
the largest descriptor set and the one that overall provides the best results.

\subsection{Methods for imbalanced data}
\label{sect:dataimbalance}

A key property of the data for all four assays is that active compounds are rare
(Table~\ref{tab:assays}),
suggesting that methods specific to imbalanced data might provide a more relevant benchmark for comparison
than off-the-shelf RF.

\citet{ChaBowHal2002} proposed the synthetic minority over-sampling technique (SMOTE)
and applied it to an assay from the National Cancer Institute's yeast anti-cancer drug screen.
The method artificially generates new examples of the minority class objects using their nearest neighbors.
In addition, the majority class is under-sampled, leading to a balanced data set.
Whereas these authors combined SMOTE with a single classification tree (C4.5),
we combine it with RF, widely recognized as more powerful for drug-discovery data.
The {\texttt R} function \texttt{SMOTE} in the package \texttt{DMwR} with default settings
adjusts the AID 348 data, for instance, in the following way.
The original $48$ active compounds are augmented with $48 \times 2 = 96$ synthetic samples.
Furthermore, the inactive compounds are randomly under-sampled to leave $96 \times 2 = 192$ cases.
Hence, the SMOTE data have a more balanced $144$ active compounds and $192$ inactive compounds.
The results for SMOTE with RF, called SRF in Tables~\ref{mean:AveP:aid348} and~\ref{tab:mean:AveP:aid362}--\ref{tab:mean:AveP:aid371},
show that overall SRF performs about the same as RF here, and sometimes worse.
EPX still dominates.

\begin{table}
\tablewidth=230pt
\caption{IE for RF, RRF and EPX averaged over $16$ balanced 10-fold cross-validations
for the AID 348 assay and five descriptor sets}
\label{tab:IE:aid348}
\begin{tabular*}{\tablewidth}{@{\extracolsep{\fill}}lccccc@{}}
\hline
\textbf{Ensemble} & \textbf{AP} & \textbf{BN} & \textbf{CAP} & \textbf{FP} & \textbf{PH}\\
\hline
RF & 5.19 & 6.62 & 7.16 & 7.07 & 5.53\\
RRF & 5.80 & 6.25 & 6.83 & 7.28 & 5.78\\
EPX & 6.27 & 8.80 & 8.20 & 8.40 & 6.44\\
\hline
\end{tabular*}
\end{table}

\begin{table}[b]
\tablewidth=\textwidth
\caption{AveP averaged over $16$ replications of balanced $10$-fold cross-validation
for EPX, RF, RRF, SRF and WRF applied to the AID 362 assay and five descriptor sets (DS).
The last four columns show the number of times EPX has better AveP than RF, RRF, SRF and WRF over the $16$ replications}
\label{tab:mean:AveP:aid362}
\begin{tabular*}{\tablewidth}{@{\extracolsep{\fill}}lcccccccccc@{}}
\hline
&  & \multicolumn{5}{c}{\textbf{Mean} \textbf{AveP}} & \multicolumn{4}{c@{}}{\textbf{EPX} \textbf{beats}} \\[-4pt]
&& \multicolumn{5}{l}{\hrulefill} & \multicolumn{4}{l@{}}{\hrulefill}\\
\textbf{DS} & \textbf{Run} & \textbf{EPX} & \textbf{RF} & \textbf{RRF} & \textbf{SRF} & \textbf{WRF} & \textbf{RF} & \textbf{RRF} & \textbf{SRF} & \textbf{WRF}\\
\hline
AP  & 1 & $0.300$ & $0.280$  & $0.256$  & $0.159$ & $0.203$  & $16/16$ & $16/16$ & $16/16$ & $16/16$\\
& 2 & $0.306$ & & & & & $15/16$ & $16/16$ & $16/16$ & $16/16$ \\
& 3 & $0.295$ &  & & & &  $13/16$ & $15/16$ & $16/16$ & $16/16$ \\[3pt]
BN  & 1 & $0.261$ & $0.242$ & $0.238$ & $0.129$ & $0.274$ & $16/16$ & $16/16$ & $16/16$ & $02/16$ \\
& 2 & $0.299$ &
 & &  & & $16/16$ & $16/16$ & $16/16$ & $15/16$\\
& 3 & $0.285$ &
  & &  & & $16/16$ & $16/16$ & $16/16$ & $14/16$ \\[3pt]
CAP  & 1 &  $0.363$ & $0.267$  & $0.171$  & $0.178$ & $0.197$  & $16/16$ & $16/16$ & $16/16$ & $16/16$ \\
& 2 & $0.355$ &
 & &  & & $16/16$ & $16/16$ & $16/16$ & $16/16$ \\
& 3 & $0.368$ &
 & &  & & $16/16$ & $16/16$ & $16/16$ & $16/16$\\[3pt]
FP  & 1 & $0.315$ & $0.266$  & $0.174$  & $0.196$  & $0.188$  & $16/16$ & $16/16$ & $16/16$ & $16/16$ \\
& 2 & $0.323$ &
 & &  & & $16/16$ & $16/16$ & $16/16$ & $16/16$ \\
& 3 & $0.306$ &
 & &  & & $16/16$ & $16/16$ & $16/16$ & $16/16$ \\[3pt]
PH  & 1 & $0.227$ & $0.216$ & $0.168$  & $0.167$  & $0.142$  & $14/16$ & $16/16$ & $16/16$ & $16/16$ \\
& 2 & $0.212$ &
 & &  & & $07/16$ & $16/16$ & $16/16$ & $16/16$ \\
& 3 & $0.218$ &
 & &  & & $09/16$ & $16/16$ & $16/16$ & $16/16$\\
 \hline
\end{tabular*}
\end{table}

\begin{table}[t]
\tablewidth=\textwidth
\caption{AveP averaged over $16$ replications of balanced $10$-fold cross-validation
for EPX, RF, RRF, SRF and WRF applied to the AID 364 assay and five descriptor sets (DS).
The last four columns show the number of times EPX has better AveP than RF, RRF, SRF and WRF over the $16$ replications}
\label{tab:mean:AveP:aid364}
\begin{tabular*}{\tablewidth}{@{\extracolsep{\fill}}lcccccccccc@{}}
\hline
&  & \multicolumn{5}{c}{\textbf{Mean} \textbf{AveP}} & \multicolumn{4}{c@{}}{\textbf{EPX} \textbf{beats}} \\[-4pt]
&& \multicolumn{5}{l}{\hrulefill} & \multicolumn{4}{l@{}}{\hrulefill}\\
\textbf{DS} & \textbf{Run} & \textbf{EPX} & \textbf{RF} & \textbf{RRF} & \textbf{SRF} & \textbf{WRF} & \textbf{RF} & \textbf{RRF} & \textbf{SRF} & \textbf{WRF}\\
\hline
AP  & 1 & $0.291$ & $0.265$  & $0.230$  & $0.204$  & $0.289$  & $16/16$ & $16/16$ & $16/16$ & $09/16$\\
& 2 & $0.292$ &
 & & & & $16/16$ & $16/16$ & $16/16$ & $09/16$ \\
& 3 & $0.310$ &
 & &  & & $16/16$ & $16/16$ & $16/16$ & $15/16$ \\[3pt]
BN  & 1 & $0.371$ & $0.327$  & $0.300$  & $0.174$  & $0.274$  & $16/16$ & $16/16$ & $16/16$ & $16/16$ \\
& 2 & $0.365$ &
 & &  & & $16/16$ & $16/16$ & $16/16$ & $16/16$\\
& 3 & $0.373$ &
  & &  & & $16/16$ & $16/16$ & $16/16$ & $16/16$ \\[3pt]
CAP  & 1 &  $0.379$ & $0.334$  & $0.252$  & $0.244$  & $0.269$  & $16/16$ & $16/16$ & $16/16$ & $16/16$ \\
& 2 & $0.390$ &
 & &  & & $16/16$ & $16/16$ & $16/16$ & $16/16$ \\
& 3 & $0.390$ &
 & &  & & $16/16$ & $16/16$ & $16/16$ & $16/16$\\[3pt]
FP  & 1 & $0.318$ & $0.305$  & $0.261$  & $0.257$  & $0.202$  & $15/16$ & $16/16$ & $16/16$ & $16/16$ \\
& 2 & $0.320$ &
 & &  & & $16/16$ & $16/16$ & $16/16$ & $16/16$ \\
& 3 & $0.317$ &
 & &  & & $14/16$ & $16/16$ & $16/16$ & $16/16$ \\[3pt]
PH  & 1 & $0.278$ & $0.275$  & $0.219$  & $0.185$  & $0.081$ & $11/16$ & $16/16$ & $16/16$ & $16/16$ \\
& 2 & $0.276$ &
 & &  & & $09/16$ & $16/16$ & $16/16$ & $16/16$ \\
& 3 & $0.282$ &
 & &  & & $14/16$ & $16/16$ & $16/16$ & $16/16$ \\
 \hline
\end{tabular*}
\end{table}

\begin{table}
\tablewidth=\textwidth
\caption{AveP averaged over $16$ replications of balanced $10$-fold cross-validation
for EPX, RF, RRF, SRF and WRF applied to the AID 371 assay and five descriptor sets (DS).
The last four columns show the number of times EPX has better AveP than RF, RRF, SRF and WRF
over the $16$ replications}
\label{tab:mean:AveP:aid371}
\begin{tabular*}{\tablewidth}{@{\extracolsep{\fill}}lcccccccccc@{}}
\hline
&  & \multicolumn{5}{c}{\textbf{Mean} \textbf{AveP}} & \multicolumn{4}{c@{}}{\textbf{EPX} \textbf{beats}} \\[-4pt]
&& \multicolumn{5}{l}{\hrulefill} & \multicolumn{4}{l@{}}{\hrulefill}\\
\textbf{DS} & \textbf{Run} & \textbf{EPX} & \textbf{RF} & \textbf{RRF} & \textbf{SRF} & \textbf{WRF} & \textbf{RF} & \textbf{RRF} & \textbf{SRF} & \textbf{WRF}\\
\hline
AP  & 1 & $0.327$ & $0.315$  & $0.281$  & $0.311$  & $0.313$  & $16/16$ & $16/16$ & $16/16$ & $16/16$\\
& 2 & $0.331$ &
 & & & & $16/16$ & $16/16$ & $16/16$ & $16/16$ \\
& 3 & $0.328$ &
 & & & & $16/16$ & $16/16$ & $16/16$ & $15/16$\\[3pt]
BN  & 1 & $0.342$ & $0.335$  & $0.333$  & $0.289$  & $0.322$  & $16/16$ & $16/16$ & $16/16$ & $16/16$ \\
& 2 & $0.354$ &
 & & & & $16/16$ & $16/16$ & $16/16$ & $16/16$\\
& 3 & $0.338$ &
 & & & & $13/16$ & $13/16$ & $16/16$ & $16/16$ \\[3pt]
CAP  & 1 &  $0.390$ & $0.347$  & $0.310$  & $0.342$  &$0.356$  & $16/16$ & $16/16$ & $16/16$ & $16/16$ \\
& 2 & $0.384$ &
 & & & & $16/16$ & $16/16$ & $16/16$ & $16/16$ \\
& 3 & $0.378$ &
 & & & & $16/16$ & $16/16$ & $16/16$ & $16/16$\\[3pt]
FP  & 1 & $0.358$ & $0.362$  & $0.338$  & $0.338$  & $0.320$  & $03/16$ & $15/16$ & $16/16$ & $16/16$ \\
& 2 & $0.358$ &
 & & & & $04/16$ & $14/16$ & $16/16$ & $16/16$ \\
& 3 & $0.364$ &
 & & & & $12/16$ & $16/16$ & $16/16$ & $16/16$ \\[3pt]
PH  & 1 & $0.277$ & $0.277$  & $0.282$  & $0.267$  & $0.244$ & $09/16$ & $05/16$ & $12/16$ & $16/16$ \\
& 2 & $0.284$ &
 & & & & $15/16$ & $10/16$ & $14/16$ & $16/16$ \\
& 3 & $0.279$ &
& & & & $09/16$ & $06/16$ & $13/16$ & $16/16$\\
\hline
\end{tabular*}
\end{table}

The method of weighted random forests [WRF, \citet{CheLiaBre2004}] was partially motivated by QSAR applications
with imbalanced data.
It assigns large and small weights to the minority and majority class compounds, respectively.
Because of a known bug with weighting in the R package \texttt{randomForest},
we increased the weight of active compounds by duplicating them.
For example, to make the AID 348 assay data balanced,
the $48$ active compounds were repeated $102$ times to have approximately as many cases
as the $4898$ inactive compounds.
The results for WRF in Tables~\ref{mean:AveP:aid348} and~\ref{tab:mean:AveP:aid362}--\ref{tab:mean:AveP:aid371}
again show little practical improvement versus RF.
For AID 362 and the BN descriptors, WRF performs better than RF and approaches
the mean AveP of one of the EPX models,
but CAP is the descriptor set of choice here,
and WRF performs worse than RF for it.

\subsection{Initial groups}
\label{sect:igroups}

For the drug-discovery application,
the initial groups presented to the phalanx formation algorithm
consist of variables in a descriptor set with related names (Section~\ref{sect:phalanx:formation}).
In other applications without logical groups of names, how should initial groups be formed?
Here we demonstrate a data-adaptive method that does not use name information.

The goal is to find a diverse set of phalanxes.
For binary descriptors, the Jaccard dissimilarity index is appropriate
and defined between binary variables $x_i$ and $x_j$ as
\[
d_{J}(x_i, x_j) =1 - \frac{x_i \cap x_j}{x_i \cup x_j}.
\]
Here $x_i \cap x_j$ is the number of observations where $x_i$ and $x_j$ both take the value~1,
and $x_i \cup x_j$ is the number of observations where $x_i$ or $x_j$ take the value 1,
and $0 \leq d_{J}(x_i, x_j) \leq 1$.
We compute the Jaccard distances between variables
via the \texttt{vegdist} function in package \texttt{vegan}
and hierarchically cluster the variables (not observations)
using \texttt{hclust} in \texttt{R} with method \texttt{ward}.
For consistency,
the number of clusters equals the number of groups formed from the names.
For example, the AP descriptors of AID 348 again have $75$ initial groups.

\begin{table}
\tablewidth=\textwidth
\caption{AveP from EPX averaged over $16$ replications of balanced $10$-fold cross-validation,
with initial groups formed from the variables' names (Names) or clusters based on the Jaccard index (Clusters)}
\label{igroups:mean:AveP:aid348}
\begin{tabular*}{\tablewidth}{@{\extracolsep{\fill}}lccccc@{}}
\hline
&  & \multicolumn{4}{c@{}}{\textbf{Mean} \textbf{AveP} \textbf{of} \textbf{EPX}} \\[-4pt]
&& \multicolumn{4}{l@{}}{\hrulefill}\\
& & \multicolumn{2}{c}{\textbf{AID} \textbf{348}} & \multicolumn{2}{c}{\textbf{AID} \textbf{371}} \\[-6pt]
&& \multicolumn{2}{l}{\hrulefill} & \multicolumn{2}{l@{}}{\hrulefill}\\
\multirow{2}{42pt}[11pt]{\textbf{Descriptor} \textbf{set}} & \textbf{Run} & \textbf{Names} & \textbf{Clusters} & \textbf{Names} & \textbf{Clusters} \\
\hline
AP  & 1 & $0.182$ & $0.200$ & $0.327$ & $0.339$\\
& 2 & $0.194$ & $0.163$ & $0.331$ & $0.334$ \\
& 3 & $0.146$ & $0.121$ & $0.328$ & $0.336$ \\[3pt]
CAP  & 1 &  $0.201$ & $0.258$ & $0.390$ & $0.386$\\
& 2 & $0.184$ & $0.275$ & $0.384$ & $0.388$\\
& 3 & $0.155$ & $0.167$ & $0.378$ & $0.380$\\[3pt]
FP  & 1 & $0.157$ & $0.166$ & $0.358$ & $0.373$\\
& 2 & $0.130$ & $0.170$ & $0.358$ & $0.364$\\
& 3 & $0.157$ & $0.175$ & $0.364$ & $0.364$\\[3pt]
PH & 1 & $0.108$ & $0.087$ & $0.277$ & $0.284$\\
& 2 & $0.108$ & $0.102$ & $0.284$ & $0.276$\\
& 3 & $0.108$ & $0.113$ & $0.279$ & $0.273$\\
\hline
\end{tabular*}
\end{table}

Table~\ref{igroups:mean:AveP:aid348} compares average AveP for EPX
with initial groups based on the variables names versus initial groups from clustering.
Results are given for assays AID 348 and AID 371 and the four binary descriptor sets.
These two assays have the smallest and largest proportions of actives, respectively,
and cover a range of performances of EPX relative to RF.
The results show that neither method for generating initial groups is uniformly better.
One difference of note is that for AID 348,
EPX with clusters makes some improvement with the CAP descriptors.
As CAP and AP already perform well for AID 348,
CAP with clusters emerges as the method of choice.
Overall, the clustering method is a viable data-adaptive approach
to forming initial groups here.

\subsection{Diversity}\label{sect:diversity}

Breiman (\citeyear{Bre2001a}) argued that the classification performance of an ensemble method
increases with the strengths of the underlying classifiers being averaged and their diversity.
We now illustrate that ensembles from phalanxes can have these desirable traits.

\begin{figure}

\includegraphics{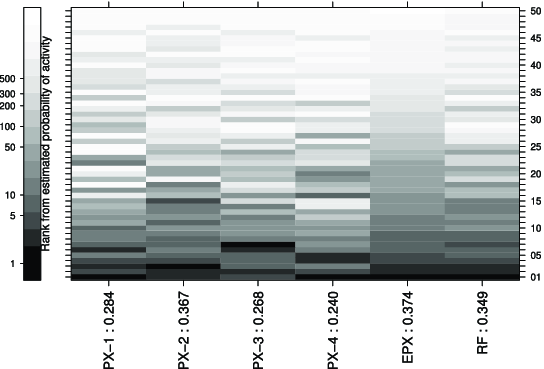}

\caption{Diversity map of ranks for the AID 364 assay and BN descriptors
from the first run of EPX and the first cross-validation.
The AveP values of PX-1 through PX-4, EPX and RF are given on the $x$-axis.
The $y$-axis has the $50$ active compounds,
ordered by their ranks from EPX.
A darker gray on the gray scale indicates a smaller rank (higher probability of activity).}
\label{fig:diversity:aid364:BN}
\end{figure}

Figure~\ref{fig:diversity:aid364:BN} depicts a diversity map [\citet{HugBroWel2012}]
of ranks from cross-validated probabilities for AID 364 and the BN descriptors.
The ranks of the $50$ active compounds are shown
for the four underlying phalanxes (denoted PX-1 through PX-4)
and their EPX ensemble in the first EPX run,
and for random forests.
This particular map relates to the first cross-validation.
The 50 active compounds on the right axis of the figure
are ordered according to the ranks from EPX.
Ideally, they would have ranks 1--50,
depicted by black to mid-gray on the gray scale on the left of the figure.
Lighter colors indicate a failure to rank well the active compounds.

It is seen in Figure~\ref{fig:diversity:aid364:BN} that PX-1 through PX-4 assign different ranks,
that is, they show a fair degree of diversity.
This is beneficial, as averaging probabilities providing the same ranks
is unlikely to improve on the underlying performances.
If one phalanx misses an active compound
(lighter gray on the rank scale in Figure~\ref{fig:diversity:aid364:BN}),
we see  that other phalanxes might rank it well
(darker gray on the rank scale).

Moreover, the AveP values for PX-1 through PX-4 reported on the $x$-axis of Figure~\ref{fig:diversity:aid364:BN}
range from 0.240 to 0.367.
PX-2 by itself beats the $0.349$ AveP value from RF using all $24$ BN descriptor variables.
Thus, PX-1 through PX-4 include one classifier that is strong relative
to RF, and they constitute a diverse set for an ensemble.
Hence, the AveP of 0.374 for EPX is also relatively high.\vadjust{\goodbreak}

AID 364 is highlighted in this analysis,
because it is a whole-cell live/dead assay likely to have multiple mechanisms of activity
from multiple chemical structures.
We now explore how diversity of the four phalanxes formed from the BN descriptors translates
into diversity of the active chemical structures identified by them.

\begin{figure}[b]\vspace*{-6pt}
\begin{tabular}{@{}ccc@{}}

\includegraphics{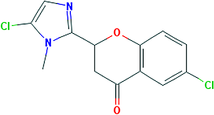}
 & \includegraphics{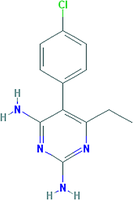} &\includegraphics{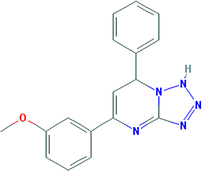} \\
\footnotesize{(a) CID 657803 (PX-1)}& \footnotesize{(b) CID 4993 (PX-1)}& \footnotesize{(c) CID 5389334 (PX-1)}\\[6pt]

\includegraphics{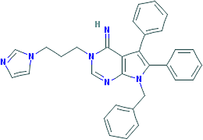}
 & \includegraphics{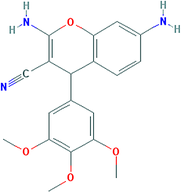} &\includegraphics{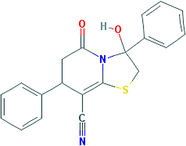} \\
\footnotesize{(d) CID 661535 (PX-2)}& \footnotesize{(e) CID 657713 (PX-1 and PX-3)} & \footnotesize{(f)  CID 661140 (PX-4)}
\end{tabular}
       \caption{Six active compounds ranked substantially higher
       by EPX than RF for assay AID 364 and the BN descriptors.
       For each compound, the phalanx(es) giving a high rank are indicated after the compound identification (CID).}
       \label{fig:six:compounds:aid364}
\end{figure}

We focus attention on the structures differentiating EPX's performance.
Sorting the active compounds by their absolute difference in ranks from EPX versus RF
leads to Table~\ref{sixactives:epx:aid364}.
\begin{table}
\tablewidth=\tablewidth
\caption{Six active compounds
where EPX and RF have the largest
absolute difference in ranks
for the AID 364 assay and BN descriptors.
The actives are identified by PubChem's compound identification (CID) number.
Ranks are also given for the individual phalanxes;
the best rank among them is in~bold}
\label{sixactives:epx:aid364}
\begin{tabular*}{\tablewidth}{@{\extracolsep{\fill}}lcccccc@{}}
  \hline
 & \multicolumn{6}{c@{}}{\textbf{Rank} \textbf{from}} \\[-4pt]
& \multicolumn{6}{l@{}}{\hrulefill}\\
\textbf{CID} & \textbf{PX-1} & \textbf{PX-2} & \textbf{PX-3} & \textbf{PX-4} & \textbf{EPX} & \textbf{RF}\\
  \hline
657713 & \phantom{0}106 & \phantom{0}715 & \phantom{0}\textbf{102} & 1082 & 133 & \phantom{0}943 \\
661140 & 2099 & \phantom{0}353 & \phantom{0}649 & \phantom{00}\textbf{52} & 158 & \phantom{0}847 \\
657803 & \phantom{0}\textbf{101} & \phantom{0}651 & 1833 & \phantom{0}744 & 241 & 1220 \\
4993 & \phantom{00}\textbf{94} & 1003 & 2673 & \phantom{0}932 & 245 & 2040 \\
5389334 & \phantom{0}\textbf{114} & 2737 & 1833 & 2376 & 364 & 2040 \\
661535 & \phantom{0}537 & \phantom{0}\textbf{211} & 1511 & 2376 & 738 & 1417 \\
   \hline
\end{tabular*}\vspace*{-3pt}
\end{table}
The six largest discrepancies all favor EPX:
there are no actives ranked substantially higher by RF than by EPX.
The structures of these six compounds presented in Figure~\ref{fig:six:compounds:aid364}
have differences in some sub-structures, particularly across phalanxes,
which may be chemically significant.
Identifying a variety of structures offers more leads for the
next stages of drug development,
where compounds are adjusted to increase efficacy,
and compounds that are toxic and mutagenic in further screens have to be removed.

The inactive compounds in the list of top 300 compounds identified by EPX
also show some diversity.
Clustering the 270 inactive compounds in the list,
according to the 24 BN descriptors  via \texttt{hclust} in \texttt{R} with method \texttt{ward},
leads to two distinct clusters according to the CH index computed by \verb!NbClust!.
Again, this is not surprising, as all the compounds in the top-ranked list,
including the inactives,
result from several phalanx models.

\begin{figure}
\begin{tabular}{@{}cc@{}}

\includegraphics{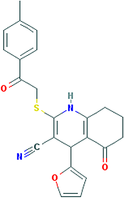}
 & \includegraphics{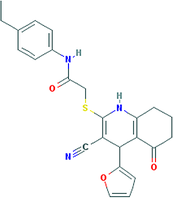}\\
\footnotesize{(a) CID 661658 (active)}& \footnotesize{(b)  CID 660076 (inactive)}
\end{tabular}
\caption{Compounds CID 661658 and CID 660076
       are ranked $3$ and $12$, respectively, by EPX for assay AID 364 and the BN descriptors.
       They have similar structures, but CID 661658 is active, while CID 660076 is not.}
      \label{fig:two:structures:aid364}
\end{figure}

Further inspection of the compounds classified by EPX as active is revealing.
Compounds CID 661658 and CID 660076 are ranked 3 and 12 by EPX and have the
similar structures shown in Figure~\ref{fig:two:structures:aid364}.
Compound CID 661658 is active, whereas CID 660076 is not.
A small difference in structure like this, that determines activity versus inactivity,
may be helpful to a chemist in designing an even more potent
structure.

\section{Conclusions and discussion}
\label{sect:conclusions}

The concept of phalanxes of variables was motivated by data sets
with little information in the response variable relative to the
dimensionality of the explanatory variables.
Scant information in the response variable
arises in drug discovery because molecules in the biologically active
class of interest are rare.
Thus, it will be difficult to use more than a few of the explanatory
variables in a single model, even if many of them are potentially useful.
An ensemble of phalanxes uses distinct explanatory variables in each
of several models,
hence, many variables have a chance to contribute to classification performance.
The phalanx-formation algorithm is guided by this aim.
The best ranking performance, both in absolute terms and relative to RF and RRF,
is seen for the CAP descriptor set, which has the most variables.
Adapting RF for imbalanced data did not make it competitive with EPX here.
Thus, we speculate that it is the sparsity of information, caused by imbalance,
and not imbalance itself, that is the key factor in EPX's performance.

Often one of the phalanxes by itself will give a classifier that outperforms
RF or RRF using all the variables.
In this sense phalanx-formation provides an effective variable selection
or regularization for QSAR studies [\citet{GooDejVan2012}].
But this is just a bonus.
The EPX algorithm attempts to identify several such competitive subsets of variables
in its various phalanxes.
Averaging their models in an ensemble usually provides the greatest advantage.

The proposed method is not a stand-alone classifier.
It works on top of a base method.
For the drug discovery problem the base classifier was RF,
because that method was among the best known for such applications.
The phalanx method divides up the variables
and gives each subset of them to the RF method.
It then combines the different results from the various RF models
(one model per subset of variables) into one model.
In that sense, the phalanx method sits on top of a base classifier like RF
and improves it.

Thus, the method is potentially extensible to other applications
with a richness of explanatory variables but other
statistical aims such as regression or survival analysis.
The user needs to provide a competitive base statistical method for the problem
and a metric for the quality of a model.
Phalanx-formation to improve the base method
would then be guided by the metric, closely following
the algorithm in Section~\ref{sect:algorithm}.

\section*{Acknowledgments}
We thank the reviewers for their insightful suggestions and comments,
which clarified important aspects of the application and methodology.
We also thank Stan Young for his help generating the descriptor sets
and Tom Pfeifer for help with chemistry interpretation. The WestGrid computational facilities of Compute Canada are also gratefully acknowledged.



\printaddresses
\end{document}